\documentclass[runningheads]{llncs}
\usepackage[T1]{fontenc}
\usepackage{cite}
\usepackage{amsmath,amssymb,amsfonts}
\usepackage{algorithmic}
\usepackage{graphicx}
\usepackage{textcomp}
\usepackage{xcolor}
\def\BibTeX{{\rm B\kern-.05em{\sc i\kern-.025em b}\kern-.08em
    T\kern-.1667em\lower.7ex\hbox{E}\kern-.125emX}}

\usepackage{ifluatex}
\usepackage[ruled,linesnumbered]{algorithm2e}

\newtheorem{Problem}{Problem}

\newcommand{\cnp}{\textbf{NP}}

\newcommand{\QED}{\hfill\rule{2mm}{2mm}}

\usepackage{xcolor}
\usepackage{enumitem}

\newcommand{\bbf}{\mbox{$\mathbb{F}$}}
\newcommand{\bbn}{\mbox{$\mathbb{O}$}}
\newcommand{\bbo}{\mbox{$\mathbb{T}$}}
\newcommand{\bbp}{\mbox{$\mathbb{P}$}}
\newcommand{\bbc}{\mbox{$\mathbb{C}$}}

\newcommand{\vecw}{\mbox{\textbf{w}}}
\newcommand{\vecx}{\mbox{\textbf{x}}}

\newcommand{\vecy}{\mbox{\textbf{y}}}

\usepackage[utf8]{inputenc} 
\usepackage[T1]{fontenc}    
\usepackage{hyperref}       
\usepackage{url}            
\usepackage{booktabs}       
\usepackage{amsfonts}       
\usepackage{nicefrac}       
\usepackage{microtype}      

\hypersetup{draft}   

\begin{document}

\title{Towards Auditing Unsupervised Learning Algorithms and Human Processes For Fairness}
\titlerunning{Auditing Unsupervised Learning Algorithms}
\authorrunning{Davidson and Ravi}

\author{Ian Davidson\inst{1} \and S. S. Ravi\inst{2}}
\institute{Department of Computer Science, University of California,
Davis, CA 95616, USA\\
{\email{indavidson@ucdavis.edu}\smallskip} \and
Biocomplexity Institute and Initiative, University of Virginia,
Charlottesville,\newline VA 22904 ~and~
Department of Computer Science, University at Albany --\newline State
University of New York, Albany, NY 12222, USA \\ 
\email{ssravi0@gmail.com}
}

\maketitle

\begin{abstract}
Existing work on fairness typically focuses on making known
machine learning algorithms fairer.
Fair variants of classification, clustering, outlier detection and
other styles of algorithms exist.  However, an understudied area
is the topic of auditing an algorithm's output to determine fairness.
Existing work has explored the two group classification problem for binary protected
status variables using standard definitions of statistical parity.
Here we build upon the area of auditing by exploring the multi-group
setting under more complex definitions of fairness.  

\keywords{Classification \and Auditing \and Fairness \and
Combinatorial Optimization \and Complexity}
\end{abstract}



\section{Introduction and Motivation}
The AI community has made tremendous progress towards making
algorithms fairer.  Fairness has been studied in the context of many
major ML tasks such as clustering,  classification,  ranking,
embedding and anomaly detection.  The area of  fairness and
ML algorithms can be divided loosely into three categories.
The first category explores \underline{pre-processing} data to make existing
algorithms fairer.  The fairlets approach of \cite{Chierichetti-etal-2017} is
perhaps the most well known example of pre-processing data so that
$k$-means and $k$-median algorithms are guaranteed to produced fair
classes (i.e., clusters).  The second category that \underline{adds fairness rules} into
algorithms is perhaps the most popular area.  Fairness rules have
been added to clustering \cite{kleindessner2019guarantees},  classification \cite{CR-2018},
outlier detection \cite{zhang2021towards} and ranking \cite{asudeh2019designing}. The third, and perhaps the
most understudied category, is \underline{post-processing} the results
of algorithms.  This work has two main sub-areas: (i) post-processing
to make the output of algorithms fairer \cite{DR-AAAI-2020} and (ii)
auditing the output of an algorithm \cite{kearns2018preventing}\cite{DR-ECAI-2020} to determine if it is fair (or
not).   
Our work falls into this second sub-area.  We view the
algorithm/human-process as dividing people into \underline{classes}
(e.g., outlier/inlier,  classes, category etc). We use the term
``group'' to refer to a protected status group, which in our work can
be a complex definition across multiple protected status variables (PSVs).

Auditing is particularly important  as it allows verification that
an algorithm's or human process's output is fair.   The latter is
particularly understudied as human processes are particularly
complex.  For example, we explore (Section~\ref{sec:exp2})  the topic
of auditing the fairness of California's 53 electoral districts along
13 protected statuses,  many of them taking multiple values.
Existing work on auditing has only studied outlier
detection \cite{DR-ECAI-2020} and classification
\cite{kearns2018preventing}; though this work is useful, it is
limited in several key ways.  Notably, it is limited to the
\underline{two class setting},  \underline{binary} protected status
and most importantly \underline{unweighted} settings as a measure
of fairness.  These settings are useful in \underline{selection}
problems such as job interviews  or \underline{decision} problems
such as predicting recidivism where decisions are binary.   However,
many settings do not match this situation.  Consider a credit card
company that divides its customer base into $k$ classes and offers
each class a \underline{different} loyalty bonus.  The classic two-class
auditing work  \cite{kearns2018preventing} does not fit this setting
and cannot be made to fit this setting by repeating it with a one
versus the rest group application.  Our second measure of fairness
(called ``utility weighted")
studies this situation, and we observe that it is possible for a set of classes
to be fair when ignoring weights but unfair when considering weights.
Finally,  consider our study in Section~\ref{sec:exp3} where we
audit news sources for fairness with respect to coverage of different
protected status individuals. There, we are interested in ensuring
equal coverage \underline{between} protected status groups and not on
a single protected status group. We study this in our third measure
of fairness (called ``pairwise equality'').
Our contributions are as follows.



\begin{enumerate}[leftmargin=*,noitemsep,topsep=0pt]
\item We formulate the search for unfairness as a
combinatorial optimization problem and establish its 
computational intractability (Theorem~\ref{thm:ab_unfairness_hard}),
leading to a test that cannot be easily side-stepped.
\item We search for three types of unfairness:
\begin{enumerate}
\item Count-based unfairness, which has been 
studied by the community as statistical parity.

\item A novel utility weighted unfairness which allows the benefit/utility
of some classes to be more than others.

\item A new pairwise unfairness which finds unfairness between two
groups (i.e., PSV combinations) of individuals.  
\end{enumerate} 
\item
For all three formulations, our methods allow finding unfairness across
multiple PSV values,  a topic rarely covered by the literature so far.

\item Our experiments consider detecting unfairness in classes
generated by algorithms as well as those created by human processes
(e.g., congressional districts of California (see Section~\ref{sec:exp2})
and news articles grouped by source media (see Section~\ref{sec:exp3})).
\end{enumerate}


\noindent
\textbf{Organization.}~ 
We begin by overviewing our method at a high level.  We then provide details of our count-based unfairness test and show
that it is computationally intractable. We extend that formulation
to a utility based setting and then to a utility based settings
that searches for unfairness over all classes. We then present
experimental results,  related work and conclude.

\section{High Level Overview of Our Approach}
\label{sec:high_level}

Our  approach to identify unfairness involves searching
for protected status variable (PSV)
combinations that are under represented  We begin with a basic formulation
that is similar to the classical count-based methods introduced by
others \cite{Chierichetti-etal-2017} and then introduce new types
of unfairness that we believe are interesting and useful.  Our work can
be seen as a framework for searching for unfairness.


\noindent
\textbf{How we detect unfairness.}
Our work  searches for over/under-represented PSV combinations
denoted by $\vecx{}$ (which represent groups of individuals). To
tie our work back to classic set cover formulations \cite{GJ-1979}
in theoretical computer science, we formulate our work as searching
for a minimum number of occurrences of a disjunction of PSVs (e.g.,
\texttt{Male} $\vee$ \texttt{Young}) that is an
\underline{over-represented} in a class compared to the other classes
(e.g., in the rest of the population).  By DeMorgan's law
\cite{Liu-1985}, this can also be seen as identifying an
\underline{under-represented} group corresponding to a conjunction
of PSVs (e.g., \texttt{Female} $\wedge$ \texttt{Elderly}). We search
across \underline{all} PSV combinations (groups of people) to find
examples of unfairness.  If \underline{no} such PSV combination is
returned, then we conclude that the division of people into classes
is fair. A domain expert can determine whether the type of unfairness
found is acceptable (or interesting), and our formulations can be
run again to explicitly avoid finding such examples of unfairness.

\smallskip
\noindent
\textbf{Types of unfairness considered.}
We formulate three types of unfairness as outlined in Table
\ref{tab:overview} 
but others are possible in our framework:

\begin{enumerate}[leftmargin=*,noitemsep,topsep=0pt]
\item \emph{Count-based.} This applies a rule similar to the traditional
definition of statistical parity \cite{kearns2018preventing}; 
it requires that the count of instances satisfying a PSV
combination $\vecx{}$ (normalized by the class size) in a class is nearly
the same as the proportion of the PSV count in the rest 
of the population. 
This definition of
fairness says that a division is unfair if any class violates this rule.

\item \emph{Utility weighted.} The above classic definition of
statistical parity assumes each class is equally important/desirable.
The credit card example discussed in the introduction does not meet
this assumption. To address it, we introduce a novel count-based
fairness that associates a utility/benefit with each class. Here,
rather than just counting how many of the group $\vecx{}$ appears
in a class, we perform a weighted count given the the utility values
for each class and compare this against a \underline{random} allocation of the
group across classes. Our optimization problem solves
for these utility values (within bounds chosen by a domain expert).

\item \emph{Pairwise equality.} Both types of fairness mentioned
above identify a \underline{single} PSV group ($\vecx{}$) that is
being treated unfairly.   Here we introduce a new type of fairness
that instead looks for unfairness between two PSV combinations
$\vecx{}$ and $\vecw{}$ (i.e.,  two groups of people).
\end{enumerate}

\begin{table}[t]
\begin{tabular}{|p{0.5in}|p{1.6in}|p{2.55in}|}
\hline
\footnotesize
\textbf{Name} & \textbf{Unfairness Detected} & \textbf{Test for Unfairness} \\ \hline
{Count}  
           & {The \underline{count} of the group $\vecx{}$ is under-represented in class $i$. 
}
           & { 
                $\exists \vecx{},i: P(\vecx{} | \neg \bbc_i) - $
                $P(\vecx{} | \bbc_i)  ~\ge~ (\beta - \alpha) = \gamma$~
                Formulation in Problem~\ref{prob:beta}.  
                Proposition~\ref{pro:p1_and_disp_imapct} in the supplement
                shows that this formulation is similar to 
                classic disparate impact calculations ($P(\vecx{} | \bbc_i) \approx P(\vecx{}$)).
               } \\ [0.2ex] \hline
Utility weighted & {The \underline{weighted count} of the group $\vecx{}$ in the current class division is under-represented compared to a \underline{random allocation} of group\newline
 members to classes.}
                 & { 
                    $\exists~U,\vecx{}:~ 
                   \,\left(\sum_{k}|\bbc_k|P(\vecx{} |
                    \bbc_k)\,U_k\right)  ~\le~$ \newline
                     $(N_{\vecx}/K)\sum_{k} U_k  - \gamma$~ 
                    {\footnotesize \textsf{s.t.}}\newline $a_k \le U_k \le b_k~~ \forall k$, 
                    where $N_{\vecx}$ is the number of instances covered by \vecx{} 
                    in the population\newline
                    (see Lemma~\ref{lem:exp_total_util} and 
                    Problem~\ref{prob:Wutility}).  
                  } \\  \hline
{Pairwise\newline Equality} & {For two groups $\vecx{}$ and $\vecw{}$,   their \underline{weighted} counts are substantially different, with $\vecx{}$ having less utility.}
                  & {$\exists~U,\vecx{},\vecw{} :$ 
                       $\sum_k \,U_k|\bbc_k| P(\vecw{}|\bbc_k) -$
                       $\sum_k \,U_k|\bbc_k| P(\vecx{}|\bbc_k)$ 
                      $ ~\ge~ \gamma$\newline
                       {\footnotesize \textsf{s.t.}}~ $a_k \le U_k \le b_k~~ \forall k$\newline  
                       {\footnotesize \textsf{and}}~ $\vecx{}^T\vecw{}$ = 0~
                       (see Problem~\ref{prob:Mutility}). 
                    } \\  [0.2ex]\hline
\end{tabular}
\smallskip
\caption{
The high level unfairness tests of a given division of instances into classes addressed by our 
combinatorial optimization problems. 
Symbols \vecx{} and \vecw{} represent subsets of PSVs.
In our formulations, $\gamma = \beta - \alpha$ is 
the disparity gap set by a domain expert.
For each type of unfairness, we have indicated the definition 
that specifies the corresponding optimization problem as a
mathematical program.} 
\label{tab:overview}
\vspace*{-0.15in}
\end{table}

\noindent
\textbf{Importance of searching across multiple PSVs.} In all
three types of unfairness, we search for combinations/groups of PSVs 
that cause unfairness.  This is critical as a set of classes maybe fair at the individual PSV level but not when considering multiple PSVs. For example,  the fraction of \texttt{Females} receiving a job offer maybe fair (equals the fraction of females in the population) as could be the case for \texttt{Married} individuals,  yet no \texttt{Females $\wedge$ Married} individuals may receive a job offer.  
Thus, in \underline{combination}, there is unfairness.

\noindent
\textbf{Importance of the hardness of our search problem.} 
Our work defines a combinatorial problem of searching for unfairness.
Suppose each person is represented by $m$ binary PSVs. Then there
are $2^m$ ``types'' or ``groups'' of people, and we must determine whether any
combination of them is treated unfairly. It is
tempting to say that such a search problem is obviously intractable; however, 
many problems with exponentially large search spaces
have polynomial time algorithms (e.g., 2SAT, the Satisfiability
problem in which each clause has at most two literals \cite{Papa-1994}).
We demonstrate the difficulty of developing efficient algorithms for our search
formulations by showing that our basic search problem (i.e., testing
for count-based unfairness) 
is computationally intractable
(Theorem~\ref{thm:ab_unfairness_hard}). This is an important property
for the following reason:
\emph{if detecting unfairness is computationally hard, it means that
making a result fairer by post processing is also computationally hard}. 
In other words, if an
algorithm produces a classification $\Pi$ into some number of
classes and our optimization formulation
finds an example of unfairness, then one cannot easily move around
a few points to obtain another classification $\Pi'$ which is fair, 
\underline{even if it is known
why $\Pi$ is unfair}!
Anecdotally,  this is because even if we know a PSV combination that makes $\Pi$
unfair, when we fix it, we may introduce other combinations that
cause unfairness.
For certain fairness measures,
this can be done efficiently in the single PSV case \cite{DR-AAAI-2020} 
but not for the case of multiple PSVs. 

\section{A Formulation for Count-Based Group Unfairness}
\label{sec:formulations}
\begin{table*}
\begin{center}
\begin{tabular}{|c|p{3.9in}|}
\hline
\textbf{Variable} & \textbf{Meaning} \\ \hline
\bbp{},~ $m$           & {The set and the number of 
                         PSVs (i.e., $m = |\bbp|$).} \\ \hline
\vecx{},~ \vecw{} & {Binary selection vectors for the PSVs for 
                    explanations using disjunctions. (Each vector
                    represents a subset of \bbp.)} \\ \hline
$\bbo{},~ \bbn{},~ \bbc_i$ & {The set of instances in a target class, 
                            other class and the $i^{th}$ class 
                            respectively. (We also use $r$ to denote $|\bbo|$
                            and $t_k$ to denote $|\bbn_k|$.)} \\ \hline
$y_k^j$,~ $z_k^j$ & {Indicator variables for the $j^{th}$ instance in 
                class $k$. 
                The value $y_k^j$~ ($z_k^j$)~ is 1 iff the $j^{th}$ instance
               in class $k$ is covered by  
               \vecx{} (\vecw{}).} \\ \hline
$U_1, U_2, \ldots, U_K$ & {Utility (benefit) values associated with
                          classes $C_1, C_2, \ldots, C_K$ respectively.}\\ \hline
$\alpha$,~ $\beta$ & {Bounds on coverage, 
                     with $\alpha < \beta$.  The value $\gamma = \beta - \alpha$ 
                     is the tolerance to unfairness.}\\ \hline
$k$,~ $K$ & {An index to classes and the total number of 
            classes respectively.}\\ \hline
$a_k$,~ $b_k$ & {Lower and upper bounds on the utility of the $k^{th}$ class,
                $1 \leq k \leq K$.} \\ \hline
\end{tabular}
\end{center}
\medskip
\caption{List of variables used in the mathematical programming formulations
developed in the paper.} 
\label{tab:notation}
\end{table*}

We first outline our test of unfairness for one class (the target
class) which is repeated $K$ times (where $K$ is the number
of classes)  with each class taking a
turn at being the target class.  It is important to understand
that our test is formulated as a search problem with the aim of
finding a simplest example of \textbf{unfairness}; if there is
no solution for this problems for all classes, this means the
classification is fair.  The notation used in the paper is summarized
in Table~\ref{tab:notation}.

\smallskip

\begin{figure}
\centering
\includegraphics[scale=0.6]{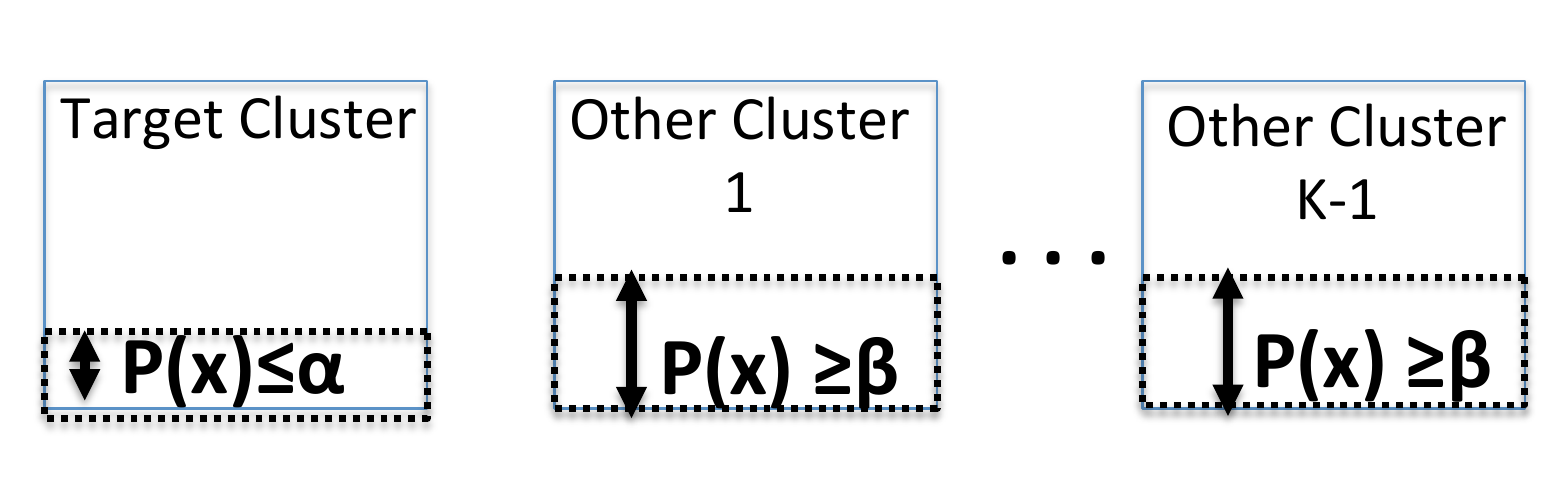}
\vspace*{-0.25cm}
\caption{A diagrammatic overview of our optimization problem
to find an explanation (denoted by \vecx{}) in terms of the PSVs
that is under-represented in one class than the others. 
The value $\gamma$ = $\beta - \alpha$
is the \emph{disparity gap}, which is a probability
for count-based unfairness and a numerical value for utility-based
unfairness.}
\label{fig:overview}
\end{figure}

\noindent 
\textbf{High-level description.} The objectives of our optimization
problems is shown 
diagrammatically in   Figure~\ref{fig:overview}.  The figure
shows $K$ Venn diagrams (one for each class), and the coverage of the
explanation (\vecx) with respect to the PSVs is 
denoted by a black dashed rectangle.  Coverage
here means that an instance $\eta$ in that class is covered by \vecx;
a formal definition of this notion of coverage is as follows.

\begin{definition}\label{def:coverage}
Let \bbc{} be a class and let vector \vecx{} represent 
a subset of (binary valued) PSVs.
The set of instances in \bbc{} \textbf{covered} by \vecx{} includes each
instance $\eta$ in \bbc{} such that at least
one PSV in \vecx{} has the value 1 in the instance $\eta$.  
\end{definition}

\noindent
\textbf{Example:}~ Suppose we have three binary PSVs, namely 
\{\texttt{Female, LowIncome, Married}\} and
\vecx{} = (1, 1, 0).
Thus, the selection vector \vecx{} represents the group/subset of individuals
given by \{\texttt{Female}~$\vee$~\texttt{LowIncome}\};
the vector \vecx{} covers any instance that represents a woman
or a person whose income is considered low (or both).

The objective of our optimization problem is to find 
a simplest\footnote{We use ``simplest'' to mean a vector $\vecx{}$
with the smallest number of PSVs.}
explanation (\vecx) such that there is a class $C_i$ where \vecx{}
is under-represented.
The extent of over (or under) representation is specified 
through a parameter $\gamma$, where $0 < \gamma < 1$, 
chosen by a domain expert which we refer to as the disparity gap.

\smallskip

\noindent
\textbf{An integer linear program (ILP) for detecting unfairness in one class.}
We now show how the unfairness detection problem mentioned above
can be expressed as an ILP.
Table~\ref{tab:notation} shows the notation used in our formulation.

Let $m = |\bbp|$ be the number of PSVs. We search for a subset 
of PSVs as given by the binary indicator vector \vecx{}.
For convenience, let $t_k = |\bbn_k|$, $1 \leq k \leq K-1$.
We compute the \emph{fraction} of instances in \bbo{}
(i.e., the target class) and $\bbn_1, \ldots,
\bbn_{K-1}$ (i.e., the other $K-1$ classes) that are covered by \vecx.  
To do this through an ILP, we represent each class $\bbn_k$ 
as an $m \times |\bbn_k|$ Boolean matrix, where each column represents 
a data point.
The column vector for the $j^{th}$  data point in $\bbn_k$,
denoted by $\bbn_k^j$, gives the 0/1 values of the $m$ PSVs for that point.
Similarly, the target class \bbo{} is considered as an $m \times |\bbo|$
matrix and its $i^{th}$ column is denoted by $\bbo^i$.

To compute the fraction of instances in \bbo{} covered by \vecx, we introduce
binary variables $z_1$, $z_2$, $\ldots$, $z_r$, where $r = |\bbo|$.
We ensure that $z_i = 1$ iff the vector \vecx{} covers the $i^{th}$
point in \bbo.
Thus, $\sum_{i=1}^{r} z_i$ gives the number of points in \bbo{} 
covered by \vecx.
We want \vecx{} to cover \underline{at most} $\alpha$ fraction of 
the points in \bbo.  

Similarly, for each class $\bbn_k$ ($1 \leq k \leq K-1$), we use
$t_k = |\bbn_k|$~ additional 0/1 variables, denoted by $y_k^1$, $y_k^2$,
$\ldots$, $y_k^{t_k}$;~ here,
variable $y_k^j$ corresponds to the $j^{th}$ point in class $\bbn_k$.
We create constraints so that $y_k^j = 1$ iff a
chosen vector \vecx{} covers the $j^{th}$ point in $\bbn_k$,
$1 \leq j \leq t_k$. Hence,
$\sum_{j=1}^{t_k} y_k^j$ gives the number of points of
$\bbn{}_k$ covered by the vector \vecx. 
We create constraints
to ensure that at least $\beta$
fraction of points in \textbf{each} of the classes 
$\bbn_1, \ldots, \bbn_{K-1}$ are covered by \vecx{}.  

If we set $\alpha=0.5\beta$ and a solution
to our optimization problem is found, it means that \vecx{} contains a
PSV combination that matches a subset of people that are 
under-represented in \bbo{} and over-represented in 
all of the classes $\bbn_1, \ldots, \bbn_{K-1}$
by a factor of 2. Conversely, if no solution is found, then no such
unfairness exists (given the requirements set by $\alpha$ and
$\beta$).  The ILP to achieve this is given below. 

\newcommand{\udsc}{\mbox{UDSC}}

\begin{Problem} \label{prob:beta}
\textbf{Unfairness Detection In a Single Class (\udsc{})  Problem.}~
Formally, a decision version of this problem can be expressed as follows: 
\begin{center}
$\exists \vecx{}: P(\vecx{} | \bbo) ~\le~ \alpha$,~ 
$                P(\vecx{} | \neg \bbo) ~\ge~ \beta$,~ $\alpha ~<~ \beta$.
\end{center}
Below, we specify an ILP formulation that focuses on finding a 
shortest explanation of unfairness.
\[
\mathrm{Objective:}~~ \mathrm{argmin}_{\vecx} ~~||\textbf{x}||
\]
satisfying the following constraints: 

\smallskip

\noindent
(1) For each class $\bbn_k$ $(1 \leq k \leq K-1)$, with $|\bbn_k| = t_k$,
the constraints are:
\begin{center}
$y_k^j$ $\leq$ $\vecx^T \, \bbn_k^j$ ~\textsf{and}~ 
$m\,y_k^j$ $\geq$  $\vecx^T \, \bbn_k^j$,~~  $1 \leq j \leq t_k$.
\end{center}

\noindent
(2) For the target class $\bbo$, the constraints are:

\begin{center}
$z_i$  $\leq$ $\vecx^T \, \bbo^i$  ~\textsf{and}~
$m\,z_i$ $\geq$  $\vecx^T \, \bbo^i$,~~   $1 \leq i \leq |\bbo|$.
\end{center}

\noindent
(3)  The set of fairness-related constraints, with $t_k = |\bbn_k|$
and $r = |\bbo|$, are: 
\begin{center}
$\sum_{j=1}^{t_k} y_k^j$ $\geq$  $\beta |\bbn_k|,$~  $1 \leq k \leq K-1$ ~\textsf{and}~
$\sum_{i=1}^{r} z_i$ $\leq$ $\alpha |\bbo|$.
\end{center}

\noindent
(4)  All the variables in \vecx{} and
all the auxiliary variables $y_k^j$ 
$(1 \leq k \leq K-1$,~ $1 \leq j \leq |\bbn_k|)$ and
$z_i$ $(1 \leq i \leq |\bbo|)$ take on values from $\mathrm{\{0,1\}}$.
\end{Problem}

\noindent
\textbf{Notes:} 

\smallskip

\begin{enumerate}[leftmargin=*,noitemsep,topsep=0pt]
\item We use $||\vecx||$ to denote the number of variables in \vecx{}
which are set to 1.
Thus, this formulation tries to find a smallest explanation
of unfairness (if one exists).
\item Let us consider the set of constraints (1) above. 
The constraint~ $y_k^j \leq \vecx^T\,\bbn_k^j$~ ensures that if \vecx{}
does not cover the $j^{th}$ instance in $\bbn_k$, the variable $y_k^j$ is
forced to be 0.
On the other hand, if 
the $j^{th}$ instance in $\bbn_k$ is covered by \vecx,
the constraint~ $m\,y_k^j \geq \vecx^T\,\bbn_k^j$~ ensures that 
$y_k^j$ is set to 1.
Similar considerations apply to the constraints specified in (2). 

\item 
The set of constraints (3) above on the summations involving
$y$ and $z$ variables have the size of the 
respective classes on the right hand side to 
ensure that $\alpha$ and $\beta$ can be interpreted as probabilities. 
\end{enumerate}

\smallskip

\section{Extensions to Utility Based Classification}
\label{sec:utility_ext}

Previously our search for unfairness merely \emph{counted} the number
of individuals to determine unfairness. This is appropriate when there are
multiple actions with the same or similar utility/benefit. But if the
utilities ($U_1, U_2, \ldots, U_K$) of being in the different classes
can vary, then there is even more opportunity for unfairness.   Classes
with different utilities arise when each group corresponding to a class 
is treated differently.
For example, a credit card company classifying customers' records may wish to
give very different benefits/rewards to each class. Our work here tries
to identify whether such rewards/utilities (within given bounds) yield
unfairness. As before, if no solution exists, then the classification is
deemed fair.  

We divide our work on this topic into two types: 
(i) utility weighted unfairness
and (ii) pairwise utility unfairness. 
For the former, we take our
previous formulation but weight each class by its utility and compare it to expected utility.
For the latter, we
create an optimization problem that attempts to find two different
PSV combinations (denoted by \vecx{} and \vecw{}) whose expected utility
difference across \textbf{all} classes is greater than a given
threshold.

\subsection{Utility Weighted Unfairness}

In the formulation for unfairness search given in Problem~\ref{prob:beta},
we implicitly gave each class/action an equal weight. 
Here we allow these weights (which
we call ``utilities'') to become part of the search problem 
for unfairness. Our formulation here
can return both an example of unfairness (denoted by \vecx{}) as
before and also the utilities of the classes that cause the
unfairness. Going back to our example with credit card customers,
bounds on these utility values (denoted by $a$ and $b$)
can be given by a domain expert 
in accordance with the
range of rewards that are say fiscally appropriate for
an organization.  Since our previous
formulation is just a special case of this version with utilities,
this formulation can
identify unfairness which cannot be detected by count-based formulations.
We present an example in Section~\ref{app:sse:unfairness_ex} of the supplement to point out
that there are classifications where the count-based approach
doesn't detect unfairness, but the utility weighted approach
reveals possible unfairness.

Our formulation now optimizes
over additional variables for the utilities ($U = \{U_1, \ldots, U_K\}$). 
To present this formulation,
we begin with a lemma that gives an expression for the expected
total utility of the instances covered by a PSV combination \vecx{}
when such instances are distributed uniformly randomly across the
$K$ classes.

\begin{lemma}\label{lem:exp_total_util}
Let $U_k$ denote the utility assigned to class $k$, $1 \leq k \leq K$.
Suppose the instances covered by a PSV combination \vecx{} are 
distributed uniformly randomly over the $K$ classes.
Then the total expected utility of the instances covered by \vecx{} is
$(N_{\vecx}/K) \sum_{k=1}^{K} U_k$, 
where $N_{\vecx}$ is the number of instances covered by \vecx{}
in the population.
\end{lemma}

\smallskip

\noindent
\textbf{Proof:}~ See supplement.

\smallskip
This above expression for the expected total utility of the instances covered
by \vecx{} was used in the second row of Table~\ref{tab:overview}.

\begin{Problem} \label{prob:Wutility}
\textbf{Utility Weighted-Unfairness Detection.}~ 
From Lemma~\ref{lem:exp_total_util} and Table~\ref{tab:overview},
the decision version
of this problem can be expressed formally as follows.

\smallskip

\noindent
\hspace*{0.1in}
$\exists~U,\vecx{}:~ \,\left(\sum_{k}|\bbc_k|P(\vecx{} |\bbc_k)\,U_k\right)  ~\le~ \alpha$,~~ 
$(N_{\vecx}/K) \sum_{k} U_k ~\ge~ \beta$,~~ $\alpha < \beta$~  \textsf{and} \\
\hspace*{0.1in}
$a_k \leq U_k \leq b_k$,~ $1 \leq k \leq K$.
\end{Problem}

\smallskip

We present an example in Section~\ref{app:sse:unfairness_ex} of
the supplement to 
show that for a given classification,
while count-based formulation (Problem~\ref{prob:beta}) 
may not reveal unfairness, our utility-based formulation
(Problem~\ref{prob:Wutility}) can reveal unfairness.

\smallskip

We now present an integer program for Problem~\ref{prob:Wutility}.
First, we specify the variables used in the formulation.
\begin{description}
\item{(a)} To be consistent with the notation used in Problem~\ref{prob:beta},
we use $\bbn_1$, $\bbn_2$, $\ldots$, $\bbn_K$ to denote the matrix
representation of the $K$ classes.
Note that the matrix representation of $\bbn_k$ is of size $m \times t_k$,
where $t_k = |\bbn_k|$, $1 \leq k \leq K$.
As before, we use $\bbn_k^j$ to denote the $j^{\mathrm{th}}$ column 
(i.e., instance) of $\bbn_k$.
We introduce $t_k$  $\{0,1\}$-valued variables $y_k^1$, $y_k^2$, $\ldots$, 
$y_k^{t_k}$ associated with $\bbn_k$, $1 \leq k \leq K$.
The significance of these variables is the same as that in Problem~\ref{prob:beta}.

\item{(b)} We use $\bbf$ to denote the matrix representation of the population.
Note that the matrix representation of $\bbf$ is of size $m \times n$,
where $n$ is the size of the population.
We use $\bbf^i$ to denote the $i^{\mathrm{th}}$ column (i.e., instance) of $\bbf$.
We introduce $n$  $\{0,1\}$-valued variables $z_1$, $z_2$, $\ldots$, 
$z_n$ associated with $\bbf$.
Variable $z_i$ is used to check whether a PSV combination \vecx{}
covers the $i^{\mathrm{th}}$ instance of the population.
(Thus, the significance of these variables is the same as that 
of the target class in Problem~\ref{prob:beta}.
Further, $\sum_{i=1}^{n} z_i$ gives the number of instances
in the population covered by \vecx.)

\item{(c)} We have variables $U_1$, $U_2$, $\ldots$, $U_K$ to
represent the utilities of the $K$ classes.
\end{description}
We are now ready to specify the objective and constraints of the
integer program for Utility-Weighted Unfairness Detection.
The $\mathrm{objective}$ is $\mathrm{argmin}_{U,\,\vecx} ~~||\textbf{x}||$
and the constraints are as follows.

\smallskip

\noindent
(1) For each class $\bbn_k$ $(1 \leq k \leq K)$, with $|\bbn_k| = t_k$,
the constraints are:
\begin{center}
$y_k^j$ $\leq$ $\vecx^T \, \bbn_k^j$ ~~\textsf{and}~~
$m\,y_k^j$  $\geq$  $\vecx^T \, \bbn_k^j,$ ~~$1 \leq j \leq t_k$.
\end{center}

These constraints ensure that the variable $y_k^j$ is set to 1 if \vecx{}
covers the $j^{\mathrm{th}}$ instance in class $\bbn_k$ and to 0 otherwise
($1 \leq j \leq t_k$ and $1 \leq k \leq K$). 

\smallskip

\noindent
(2) For the population \bbf{}, the constraints are:
\begin{center}
$z_i$  $\leq$  $\vecx^T \, \bbf^i,$ ~~\textsf{and}~~ 
$m\,z_i$ $\geq$  $\vecx^T \, \bbf^i,$~~  $1 \leq i \leq n$.
\end{center}

These constraints ensure that the variable $z_i$ is set to 1 if \vecx{}
covers the $i^{\mathrm{th}}$ instance in the population \bbf{} and to 0 otherwise
($1 \leq i \leq n$). 

\smallskip

\noindent
(3)  The set of fairness-related constraints, with $t_k = |\bbn_k|$
and $n$ being the size of the population are:
\begin{center}
$\sum_{k=1}^{K}\left(U_k\,\sum_{j=1}^{t_k} y_k^j\right)$ 
        $\leq$   $\alpha$ ~~\textsf{and}~~
$(\sum_{i=1}^{n} z_i) \times (\sum_{k=1}^K U_k)/K$
       $\geq$  $\beta$.
\end{center}

The first constraint above uses the total utility of the instances
covered by \vecx{} in the given classification. 
The second constraint above uses the expected total utility of the instances
covered by \vecx{} in the population when these instances are distributed
randomly over the $K$ classes. (This constraint uses Lemma~\ref{lem:exp_total_util}.)

\smallskip

\noindent
(4) Bounds on utility values: $a_k \:\le\: U_k \:\le\: b_k$,~~ $1 \leq k \leq K$. 

\smallskip

\noindent
(5)  All the variables in \vecx{} and
all the auxiliary variables $y_k^j$ 
$(1 \leq k \leq K$,~ $1 \leq j \leq |\bbn_k|)$ and
$z_i$ $(1 \leq i \leq n)$ take on values from $\mathrm{\{0,1\}}$.

\smallskip

\noindent
\textbf{Note:}~
As this is a more complex search problem, 
the formulation uses non-linear constraints.
In particular, constraints in (3) above are non-linear. 
As before, the values of $\alpha$ and $\beta$ are chosen
by a domain expert depending on the desired disparity gap $\gamma$.

\subsection{Pairwise Utility Unfairness}

Here we explore the extension of our earlier formulations to
allow aggregation across multiple classes. Instead of testing whether there
exists a subset of people (denoted again by \vecx{}) who are
under-represented in one class compared to the rest, we search for
two groups of people, denoted by \vecx{}~and \vecw{}, whose expected
utility when summed up over all classes differs by a value that is
at or beyond a specified tolerance level.
allowance. 

To achieve this, we use variables $y_k^j$ for \vecx{} (and $z_k^j$
for \vecw{}) to encode whether instance $j$ in class $k$ is covered
by \vecx{} (\vecw{}). 
These indicator variables are then summed and
multiplied by the utility of each class and a constraint is imposed on
the difference that is not tolerable using a chosen disparity
threshold $\gamma$. (Recall that our optimization
problems are tests of \emph{unfairness}.) 
To achieve this, $\alpha_k$ and $\beta_k$ are the utility of instances
in class $\bbc_k$ covered by \vecx{} and \vecw{} respectively.
The final constraint places a lower bound $\gamma$ on the sum of their difference.

\begin{Problem} \label{prob:Mutility}
(\textbf{Pairwise Utility Unfairness Detection.})~
A formal statement of the decision version of this 
problem is as follows:

\smallskip

\noindent
\hspace*{0.1in}
$\exists\,U,\vecx{},\vecw{}\,:$ $\sum_k \,U_k\,|\bbc_k|\,P(\vecx{}\,|\,\bbc_k) -$ 
                       $\sum_k \,U_k\,|\bbc_k|\, P(\vecw{}\,|\,\bbc_k)$
                       $\ge$ $(\beta - \alpha) = \gamma$~
                       {\small \textsf{s.t.}}\\
\hspace*{0.1in}
$a_k \le U_k \le b_k~~ \forall k$~
                       {\small \textsf{and}}~ $\vecx{}^T\vecw{} = 0$. 


\smallskip

\noindent
An integer program for finding a shortest explanation
of unfairness is as follows.
The objective here is
$\mathrm{argmax}_{U,\,\vecx,\,\vecw} ~~||\textbf{x}-\textbf{w}||$ 
and the constraints are as follows. 

\smallskip

\noindent
(i) For each class $\bbc_k$ $(1 \leq k \leq K)$, with $|\bbc_k| = t_k$,
the constraints are as follows. $($As before, the notation $\bbc_k^j$ 
represents the $j^{th}$
column of the $m \times t_k$ Boolean matrix representing $\bbc_k$.$)$
\begin{center}
~~~~$y_k^j$  $\leq$  $\vecx^T\,\bbc_k^j$ ~~\textsf{and}~~
$m\,y_k^j$ $\geq$  $\vecx^T\,\bbc_k^j,$~~ $1 \leq j \leq t_k$ \\ [0.5ex]
~~~~$z_k^j$   $\leq$  $\vecw^T\,\bbc_k^j$ ~~\textsf{and}~~
$m\,z_k^j$ $\geq$ $\vecw^T\,\bbc_k^j,$~~ $1 \leq j \leq t_k$ \\ [0.5ex]
$\alpha_k$ $=$  $U_k\sum_{j=1}^{t_k} y_k^j$, ~~
$\beta_k$  $=$  $U_k\sum_{j=1}^{t_k} z_k^j$ 
\end{center}

\smallskip

\noindent
(ii) Other constraints: 

\smallskip

\noindent
$\sum_{k=1}^{K} \alpha_k -$ $\sum_{k=1}^{K} \beta_k$ $~\ge~ \gamma$,~
$a_k ~\le~ U_k ~\le~ b_k$~  $(1 \leq k \leq K)$~~ 
                {\small \textsf{and}}~ 
$\vecx{}^T\vecw{} = 0$.
\end{Problem}

\noindent
\textbf{Note:}~ The constraint $\vecx^T\vecw{} = 0$
above ensures that the sets of PSVs
represented by \vecx{} and \vecw{} are disjoint.
(For example, this prevents the possibility of a subset relationship
between \vecx{} and \vecw.)

\section{Proof of Computational Intractability}\label{sec:complexity}

\newcommand{\dugc}{\mbox{\textsc{UDSC}}}

This section can be skipped on first reading with the understanding
that the underlying problem of searching for the simplest count based fairness
is computationally intractable. That is, 
under a standard hypothesis in computational complexity \cite{Papa-1994},
there can be no general purpose algorithm
that finds \vecx{} efficiently. 
This is important as it points out the difficulty of efficiently
modifying an existing unfair classification to create a 
classification that is fair. 

To investigate the computational complexity of the \dugc{}
problem (defined as Problem~\ref{prob:beta}),
we use the following decision version of the problem.

\smallskip

\noindent
\textbf{Unfairness Detection in a Single Class}~ (\dugc)

\smallskip

\noindent
\underline{Given:}~ A collection of $K \geq 2$ pairwise disjoint classes 
$\bbo$, $\bbn_1$, $\ldots$, $\bbn_{K-1}$ and a set 
\bbp{} = $\{p_1, p_2, \ldots, p_m\}$
of $m$ PSVs, positive integers $\alpha$ and $\beta$, where $\alpha < \beta$.

\smallskip

\noindent
\underline{Question:}~ Is there a subset $P' \subseteq \bbp$
such that $P'$ covers at most $\alpha$ instances of $\bbo$ and
at least $\beta$ instances in each of the other classes 
$\bbn_1$, $\bbn_2$, $\ldots$, $\bbn_{K-1}$?

For simplicity in presenting the proof, we have used $\alpha$ and $\beta$
as integers in the above formulation.
It is straightforward to express them as fractions of
the population size.
Unlike the ILP formulation, \dugc{} defined above
is a decision problem; it does not require the 
minimizing the number of PSVs used in the explanation.
Nevertheless, we have the following theorem.

\begin{theorem}\label{thm:ab_unfairness_hard}
The \dugc{} problem is \cnp-complete even for two classes.
\end{theorem}

\smallskip
  
\noindent
\textbf{Proof:}~ See supplement.

\section{Experiments}
\label{sec:practice}
We explore our three formulations to measure fairness from three
different situations (clustering, human processes and classification). These serve to validate our formulations and
also illustrate their use in practical situations.

\begin{enumerate}[leftmargin=*,noitemsep,topsep=0pt]
\item \textbf{Count-Based Group Unfairness.} We evaluate
the fairness of solutions produced by existing fair-by-design 
clustering algorithms.   We observe not unexpectedly that focusing 
on a single PSV can produce unfairness with respect to other PSVs. 
This is a simple but necessary result to show the need for fairness across multiple PSVs.
\item \textbf{Utility Weighted Unfairness.} 
Here we search for examples of unfairness in the 53 congressional
districts in California amongst multiple PSVs collected during the 2010 census. 
This is an example of identifying unfairness in a historical 
classes produced by humans.
\item \textbf{Pairwise Utility Group Fairness.} We explore a novel use of 
budgeting time to read articles from multiple sources so as not to get
a biased perspective on a topic.  These sources are created by a complex decision/classification process. 
\end{enumerate}

\subsection{Evaluating the Unfairness of
Fair-By-Design Clustering Algorithms}
\label{sec:exp1}
We take the output of a classic (fairlet-based) fair-by-design clustering
algorithm \cite{Backurs2019} which ensures fairness
for just one PSV and then measure fairness across the remaining
PSVs.  Even though this is a simple experiment, we believe that
it is necessary.
We take the classic Adult Data set \cite{Dua:2019} studied by many
fair clustering papers  \cite{Chierichetti-etal-2017,Bera-etal-2019,DR-AAAI-2020,Schmidt-etal-2018,kleindessner2019fair} 
which contains four PSVs (\texttt{gender, education, marital-status, occupation}). We produce
a fair clustering for just a single PSV (as the fairlets method
allows) and then measure unfairness across the remaining three PSVs. 
In all experiments we use $K=6$ as is typical with this data set.
This is achieved by solving Problem~\ref{prob:beta} for each cluster
in turn as the target, and if any solution is returned, the clustering
is deemed unfair and the PSV combination causing the unfairness
noted. If a PSV combination is found, we re-run the formulation in
Problem~\ref{prob:beta}
again with an additional orthogonality constraint to find a new PSV
combination (example of unfairness) until no unfairness is discovered.

We set $\gamma$ to be 20\% less than the median
population probability (mean of two middle values) of all PSV
combinations. 
The results shown in Table~\ref{tab:exp1} indicate the need for measuring
unfairness across multiple PSVs. 

\begin{table*}
\begin{center}
\begin{tabular}{|c||p{3.7in}|}
\hline
\textbf{PSV Balanced} & {~\textbf{No. of Unfair Combinations 
                        in the Remaining PSVs}}  \\ \hline
Gender (\texttt{G})         &  {~5 (\texttt{E, EM, EMO, O, OM})} \\
Education (\texttt{E})      &  {~3 (\texttt{GO, GM, GMO})}\\
Marital Status (\texttt{M}) &  {~5 (\texttt{E, EO, G, GO, EGO})}\\ 
Occupation (\texttt{O})     &  {~3 (\texttt{EM,MG,EMG})} \\ \hline
\end{tabular}
\medskip
\end{center}
\caption{Measuring the fairness of the output of classic
fairness-by-design clustering algorithms \cite{Backurs2019} 
on the census/adult data set\cite{Dua:2019}. The
algorithm balanced the PSV in the left column and we report the
number and examples of unfairness found on the remaining three PSVs (maximum of 8).
Unfairness is reported if there exists a clustering which contains 
an individual that is under-represented so as to cause disparate 
impact (20\% discrepancy).
}
\label{tab:exp1}
\vspace*{-0.25in}
\end{table*}

\subsection{Evaluating Utility Based Unfairness for Census Data}
\label{sec:exp2}

The previous experiment inherently identified unfairness in a
particular class by identifying if a group of individuals was
greatly under-represented in one particular class compared to the remaining classes. 
However,  such a fairness test ignores the utility of the classes
as discussed in Section~\ref{sec:utility_ext}.
Indeed it is possible our previous test can say a solution is fair
but a utility weighted test say the opposite. (As mentioned
earlier, an example to illustrate this appears in 
Section~\ref{app:sse:unfairness_ex}.)
Here, we consider the utility of the clusters
($U_1, \ldots, U_K$) when detecting unfairness.  If a protected
status (denoted by \vecx{}) group's weighted utility for the given
set of classes is substantially different from the expected
utility (over randomly created classes) then the classification
is deemed unfair.

California consists of 53 congressional districts (CDs). 
Each of them can be considered a class containing a subset of the 1700+ Zip Code Tabulation
Areas (ZCTAs) \cite{census} as shown in Figure~\ref{fig:CD} in the supplement (Section~\ref{app:sse:cong_dist}).
For each ZCTA,  we have its assignment to a CD,  population size and the fraction of its population having the following well known demographic attributes \cite{grubesic2006use}:

\begin{center}
{\small
\begin{tabular}{lllll}
\textsf{Foreign born,} & \textsf{Chinese,} & \textsf{Black,} 
                       & \textsf{Indian,}  & \textsf{Vietnamese,}\\
\textsf{Filipino,}     & \textsf{White,}   & \textsf{65 years+,}~    
                       & \textsf{Female,}~  & \textsf{Japanese,} \\
\textsf{American Indian,}~ & \textsf{Native Hawaiian,}~ & \textsf{Islander}  & &  \\
\end{tabular}
}
\end{center}

\begin{figure}
\begin{center}
\includegraphics[scale=0.4]{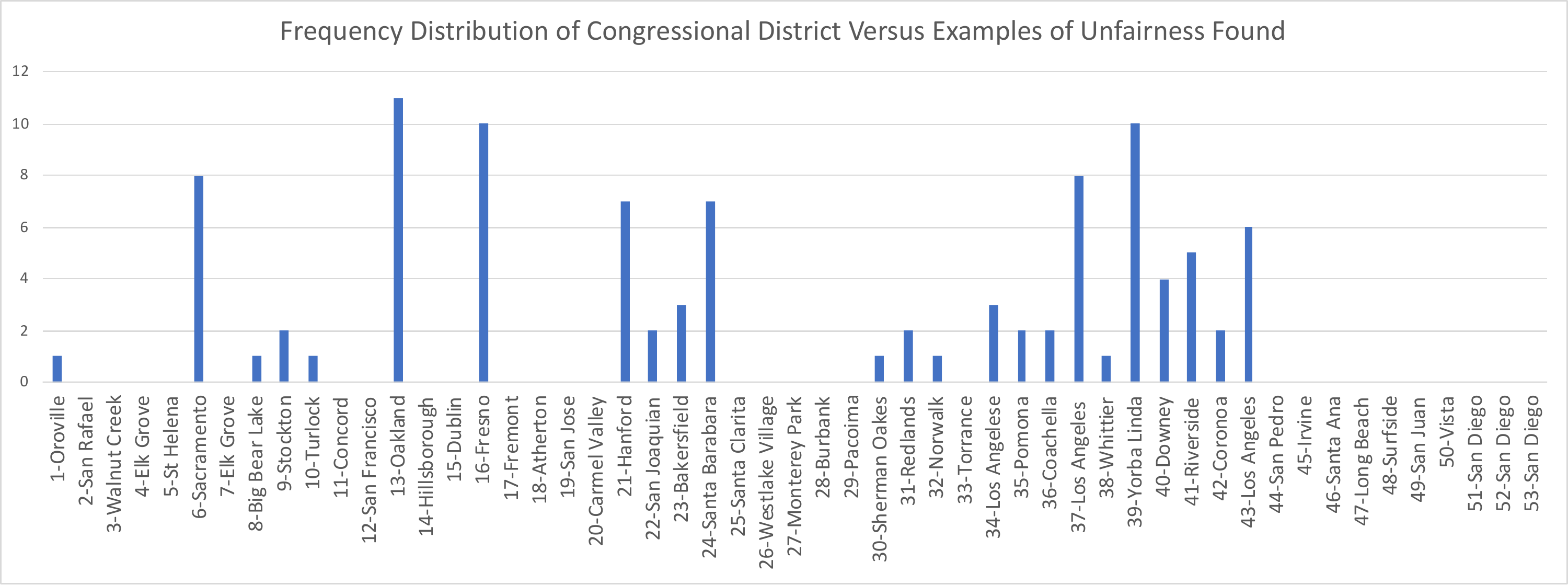}
\end{center}
\caption{
The distribution of the 100 shortest explanations/examples of
unfairness across California's 53 congressional districts. The
x-axis refers to the congressional district and the y-axis indicates
how often unfairness was found in the district.}
\label{fig:census}
\end{figure}

\noindent
We  use this information to create a
synthetic population of individuals who match the demographic
information in each CD and then measure the fairness
of the 53 CDs (classes).  Each CD has a different median local property tax basis (per capita) which is used as the utility measure as it indicates a general quality of living given local taxes fund schools, local sports, parks and other important quality of living indicators.
We use the formulation specified as Problem~\ref{prob:Wutility}.  
If no solution is found for any of these problems, then
the CDs are ``fair'' in that no PSV-combination defined group of
people  is allocated 20\% less money than their expected utility if they were assigned randomly to the CDs. 
Our method
discovers the simplest forms of unfairness and we repeat our
experiment 100 times, each time adding an orthogonality constraint
to not discover a previously found form of unfairness. We calculated
the distribution of unfairness found in the 53 CDs and found that
it is concentrated in the following districts: 13th-Oakland,   16th Fresno, 
21st Hanford  24th-Santa Barbara,
37th-Los~Angeles and 39th La Habra (see 
Figure~\ref{fig:census}).  An overwhelming fraction of the unfairness
explanations centered on race but \underline{not} on country of
birth or gender.

\subsection{Using Pairwise Utility-Based Fairness For Reading Times}
\label{sec:exp3}
Here we explore the situation of finding fairness \underline{between} different protected status groups.  	This allows finding a new style of \underline{comparative} unfairness in that group $\vecx{}$ is being given unfair (under-represented) treatment compared to another group 
$\vecw{}$.  Consider the situation where you have a collection of $r$ sources of documents with each document each on  $k$ different topics. 

We use the Twitter Dataset of Health News \cite{karami2018fuzzy}
(the topics being the health of various types of individuals) 
which contains the classified Twitter feeds of 
the following 16  health news sites/sources.  

\begin{center}
{\small
\begin{tabular}{llllll}
\textsf{bbchealth}, & \textsf{cbchealth}, & \textsf{cnnhealth}, &
\textsf{everydayhealth},  &
\textsf{foxhealth},~ & \textsf{gdnhealth}, \\
 \textsf{goodhealth},~ &
\textsf{KaiserHealth}, &
\textsf{latimeshealth},~ & \textsf{msnhealthnews},~ & \textsf{NBChealth},~ & 
\textsf{nprhealth},\\ 
\textsf{nytimeshealth},~ & \textsf{reuters-health},~ & \textsf{usnewshealth},~ &
\textsf{wsjhealth}\\ 
\end{tabular}
}
\end{center}

\noindent
An article may be on one or more of the following protected status topics 
\{\texttt{Gender, Handicapped, Poverty}\} and our aim is to get a balanced overview of each.
Each news site contains many articles (see Table~\ref{tab:exfeeds} in the supplementary material for an example).

Our third optimization formulation
(Problem~\ref{prob:Mutility}) can be used to search for a pair of
under/over represented PSV combinations. The lower ($a_1, \ldots,
a_k$) and  upper ($b_1, \ldots, b_k$) bounds on the utility values
$U_1, \ldots, U_k$ can be set as the
allowable time to spend on each news source. If no solution is found,
then we can spend between $a_k$ and $b_k$ units of time 
on news source $k$ ($1 \leq k \leq 16$) and get a balanced
(fair) view of the overall topic. Conversely, if our formulation returns a
solution, then we get over/under represented PSVs given by \vecx{} and
\vecw{}. We make two simplifying assumptions: each
article in a source is randomly chosen and all articles
take equal time to read.

In our experiment, we assume we have a total of 16 hours (i.e., 60
minutes per source) and  set $a_k$ = 50 minutes and
$b_k$ = 70 minutes, for $1 \leq k \leq 16$; more sophisticated bounds can
be set depending on the size of each repository. We set $\gamma$
to be 15, indicating that the time difference spent reading about
any two PSV combination should not be greater than 15 minutes.
After solving Problem \ref{prob:Mutility} with the above 
parameters we find our optimization problem returned no solution. 
Hence, we conclude that spending between 50 and 70 minutes per news source
won't lead to a biased account of the healthcare topic, given the
simplifying assumptions made earlier. If the problem had returned
a solution, then \vecx{} and \vecw{} identify a pair 
of under/over represented
PSV combinations (health topics).

\section{Related Work}
We discuss two areas of related work and discuss how they differ
from our own work. The first of these areas is the work on 
fairness in classification and
the second area is that of auditing classification algorithms.

\smallskip

\noindent
\textbf{Fair-by-Design Clustering/Classification.} The fairness-by-design
clustering/classification algorithms (e.g., \cite{Chierichetti-etal-2017}) measure fairness
by calculating the balance of class $i$ defined by $B_i =
\min(\frac{\#Red_i}{\#Blue_i},\frac{\#Blue_i}{\#Red_i})$, where
$\#Red_i$ ($\#Blue_i$) indicates the number of red (blue)
instances in a class.
(One can think of red and blue instances in a class as representing women and 
men respectively.) The fairness of a classification is
then simply the minimal balance across the classes, that is, 
$\min(B_1, \ldots, B_K)$. 
Optimizing this criterion is equivalent to requiring 
$P(Red\,|\,C_i) \approx P(Red)~ \forall i$, that is,
the probability of finding a red
instance in a class is equal the probability of finding a red
instance in the population; a similar condition holds
for blue instances as well \cite{Backurs2019}. 

Our work is
fundamentally different in that: (i) we are testing for fairness
where as this work generates fair classification, (ii) our tests involve
multiple PSVs and iii) our work extends beyond simple 
count-based fairness.

\smallskip

\noindent
\textbf{Auditing Classifiers.} The work on auditing classifiers
\cite{kearns2018preventing} considers the application of a binary
classifier to a data set and certify it is fair. The authors define
fairness here with respect to two properties: (i) statistical parity
and (ii) false positive group level fairness. Since our work is in
the unsupervised setting, the second property is not applicable.
To discuss the first property, we note that 
Kearns et al. \cite{kearns2018preventing} allow a user to specify
groups of instances and require statistical parity to hold for each group.
Let \emph{Red} denote one of the groups and let the target class 
be denoted by $C_+$. 
Then the statistical parity property can be viewed as 
requiring that $P(Red,C_+) = P(Red).P(C_+)$. 
This condition can be seen to imply (through simple algebraic manipulations)
that $P(Red\,|\,C_+) = P(Red)$; hence, 
the condition is equivalent to the fair-by-design criterion
for fairness discussed earlier.  
Proposition \ref{pro:p1_and_disp_imapct} (in the supplement)
shows our count based formulation is equivalent to this measure of fairness. 
However, their work is for binary classification and
focuses only on one class (the target class).  Only our count
based measure of fairness is related to this work.  
Our fairness measure requires that $\forall i$,~
$P(Red \,|\, C_i) ~\approx~ P(Red \,|\, \neg C_i)$, which for
many classes of equal size simplifies to $P(Red \,|\, C_i) \approx
P(Red)$~ $\forall i$, as $\neg C_i$ is nearly the population of instances.
However, there are significant differences. Firstly, we measure
fairness across all classes, not just one; most importantly, we
check for fairness across all possible PSV combinations  and not just for a given
set of groups as in \cite{kearns2018preventing}. 
To the best of our knowledge, our work on
utility-based fairness has not been studied in the literature.

\section{Summary and Conclusions}

Most work on fairness focuses on fair-by-design algorithms to produce
fair output.  Here, we take the alternative direction of testing
whether the output of an algorithm is fair.  We explore the topic
of testing whether a given set of classes is unfair (given parameters
set by domain experts) as a series of combinatorial optimization problems
designed to search for unfairness.

Our first formulation tested for unfairness using a count-based
definition of fairness which is similar to those measures for
statistical parity although it measures fairness across multiple
PSVs (see Proposition \ref{pro:p1_and_disp_imapct}). However, these count based methods equate unfairness with
under-representation in one class and hence inherently assume that
being in one class is equally desirable as being in another.
Using utilities to model the benefit of being in different classes
allows the search for cost-sensitive unfairness across multiple
classes which has not been studied in the fair classification literature.
Our final formulation explores the important
topic of finding pairs of protected status groups that are not being
treated equally. This is often how fairness is evaluated in challenging
situations such as access to gifted and talented education (GATE) programs
in schools. 

If no solution exists to our optimization problems we deem the
classification fair; otherwise, our methods return an explanation for
why the classification is unfair. When a solution exists, the domain
expert can determine if it is significant.  Since our formulations
lead to \cnp-hard problems, they cannot be easily side-stepped.
This means that even if we say a classification is unfair and why it is
unfair, an efficient algorithm to manipulate the existing classification
to make it fair cannot exist under a
standard hypothesis in computational complexity.  

To demonstrate the usefulness of our formulations,
we explored several new domains
including testing for fairness in California's 53 congressional
districts and how to budget time across multiple reading sources
(the classes) so as to obtain a non-biased (fair) view of a topic.

\medskip

\noindent
\textbf{Acknowledgments:}~
This work was supported in part by NSF Grants IIS-1908530 and IIS-1910306 titled:
``Explaining Unsupervised Learning: Combinatorial Optimization Formulations,
Methods and Applications''.

\bibliographystyle{spmpsci}      
\bibliography{ecai,refs}

\begin{thebibliography}{10}
\providecommand{\url}[1]{{#1}}
\providecommand{\urlprefix}{URL }
\expandafter\ifx\csname urlstyle\endcsname\relax
  \providecommand{\doi}[1]{DOI~\discretionary{}{}{}#1}\else
  \providecommand{\doi}{DOI~\discretionary{}{}{}\begingroup
  \urlstyle{rm}\Url}\fi

\bibitem{asudeh2019designing}
Asudeh, A., Jagadish, H., Stoyanovich, J., Das, G.: Designing fair ranking
  schemes.
\newblock In: Proceedings of the 2019 International Conference on Management of
  Data, pp. 1259--1276 (2019)

\bibitem{Backurs2019}
Backurs, A., Indyk, P., Onak, K., Schieber, B., Vakilian, A., Wagner, T.:
  Scalable fair clustering.
\newblock In: Proc. ICML, pp. 405--413 (2019)

\bibitem{Bera-etal-2019}
Bera, S.K., Chakrabarty, D., Negahbani, M.: Fair algorithms for clustering.
\newblock CoRR \textbf{abs/1901.02393v1} (2019)

\bibitem{Chierichetti-etal-2017}
Chierichetti, F., Kumar, R., Lattanzi, S., Vassilvitskii, S.: Fair clustering
  through fairlets.
\newblock In: Proc. NeurIPS, pp. 5036--5044 (2017)

\bibitem{CR-2018}
Chouldechova, A., Roth, A.: The frontiers of fairness in machine learning.
\newblock ArXiv: 1810.08810v1 (2018)

\bibitem{DR-ECAI-2020}
Davidson, I., Ravi, S.S.: A framework for determining the fairness of outlier
  detection.
\newblock In: Proc. ECAI 2020, pp. 2465--2472. IOS Press (2020)

\bibitem{DR-AAAI-2020}
Davidson, I., Ravi, S.S.: Making existing clusterings fairer: Algorithms,
  complexity results and insights.
\newblock In: Proc. AAAI 2020, pp. 3733--3740 (2020)

\bibitem{Dua:2019}
Dua, D., Graff, C.: {UCI} machine learning repository (2017).
\newblock \urlprefix\url{http://archive.ics.uci.edu/ml}

\bibitem{GJ-1979}
Garey, M.R., Johnson, D.S.: Computers and Intractability: A Guide to the Theory
  of {NP}-Completeness.
\newblock W. H. Freeman \& Co., San Francisco (1979)

\bibitem{grubesic2006use}
Grubesic, T.H., Matisziw, T.C.: On the use of {ZIP} code tabulation areas
  {(ZCTAs)} for the spatial analysis of epidemiological data.
\newblock International journal of health geographics \textbf{5}(1), 58 (2006)

\bibitem{karami2018fuzzy}
Karami, A., Gangopadhyay, A., Zhou, B., Kharrazi, H.: Fuzzy approach topic
  discovery in health and medical corpora.
\newblock Int. J. Fuzzy Systems \textbf{20}(4), 1334--1345 (2018)

\bibitem{kearns2018preventing}
Kearns, M., Neel, S., Roth, A., Wu, Z.S.: Preventing fairness gerrymandering:
  Auditing and learning for subgroup fairness.
\newblock In: Proc. ICML, pp. 2564--2572 (2018)

\bibitem{kleindessner2019fair}
Kleindessner, M., Awasthi, P., Morgenstern, J.: Fair k-center clustering for
  data summarization.
\newblock In: Proc. ICML, pp. 3448--3457 (2019)

\bibitem{kleindessner2019guarantees}
Kleindessner, M., Samadi, S., Awasthi, P., Morgenstern, J.: Guarantees for
  spectral clustering with fairness constraints.
\newblock Proc. ICML pp. 3458--3467 (2019)

\bibitem{Liu-1985}
Liu, C.L.: Elements of Discrete Mathematics.
\newblock McGraw-Hill, New York, NY (1985)

\bibitem{MU-2005}
Mitzenmacher, M., Upfal, E.: Probability and Compting: Randomization and
  Probabilistic Techniques in Algorithms and Data Analysis.
\newblock Cambridge University Press, New York, NY (2005)

\bibitem{Papa-1994}
Papadimitriou, C.H.: Computational Complexity.
\newblock Addison Wesley, Reading, MA (1994)

\bibitem{Schmidt-etal-2018}
Schmidt, M., Schwiegelshohn, C., Sholer, C.: Fair coresets and streaming
  algorithms for fair $k$-means clustering.
\newblock CoRR \textbf{abs/1812.10854v1} (2018)

\bibitem{census}
{US~Census~Data}.
\newblock \url{http: //www.census.gov/tiger/tms/gazetteer/zcta5.txt} (2010)

\bibitem{zhang2021towards}
Zhang, H., Davidson, I.: Towards fair deep anomaly detection.
\newblock In: Proc. FAccT, pp. 138--148 (2021)

\end{thebibliography}

\clearpage

\appendix

\medskip

\begin{center}
\fbox{{\Large\textbf{Supplementary Material}}}
\end{center}

\medskip

\section{Additional Material for Section~\ref{sec:high_level}}
\label{app:sec:high_level}

\medskip
 
We mentioned in Table~\ref{tab:overview} of Section~\ref{sec:high_level}
that our count-based
unfairness is similar to the classic disparate impact
calculation. Here, we provide a formal statement
and proof of that statement.

\medskip

\begin{proposition}\label{pro:p1_and_disp_imapct}
Suppose a set $S$ of $n$ instances is partitioned into $K$ nonempty
classes $\bbc_1$, $\bbc_2$, $\ldots$, $\bbc_K$.
Further, suppose for a PSV combination \vecx{}, 
$P(\vecx\,|\,\bbc_i)$ =  $P(\vecx\,|\,\neg\bbc_i)$ for 
each $i$, $1 \leq i \leq K$.
Then $P(\vecx\,|\,\bbc_i)$ =  $P(\vecx)$ for each $i$,
$1 \leq i \leq K$.
\end{proposition}

\medskip

\noindent
\textbf{Proof:}~ Let $N({\vecx})$ denote total number of instances
of $S$ covered by the PSV combination \vecx.
Consider any class $\bbc_i$ and
let $N(\bbc_i, \vecx)$ denote the number of instances 
of $\bbc_i$ covered by \vecx.
Thus, 
\begin{equation}\label{eqn:two_cond}
P(\vecx) ~=~ N(\vecx)/n ~~\mathrm{and}~~ 
P(\vecx\,|\,\bbc_i) ~=~ N(\bbc_i, \vecx)/|\bbc_i|.
\end{equation}
Now, we use the condition that
$P(\vecx\,|\,\bbc_i)$ =  $P(\vecx\,|\,\neg\bbc_i)$. 
Note that 
\[
P(\vecx\,|\,\neg\bbc_i) ~=~  [N(\vecx) - N(\bbc_i,\vecx)]/(n-|\bbc_i|). 
\]
Thus, the condition
$P(\vecx\,|\,\bbc_i)$ =  $P(\vecx\,|\,\neg\bbc_i)$ yields
\begin{equation}\label{eqn:inter}
N(\bbc_i,\vecx)/|\bbc_i| ~=~ [N(\vecx) - N(\bbc_i,\vecx)]/[n-|\bbc_i|].
\end{equation}
Simplifying Equation~(\ref{eqn:inter}), we get
\begin{equation}\label{eqn:req_cond}
N(\vecx)/n ~=~ N(\bbc_i, \vecx)/|\bbc_i|.
\end{equation}
By inspecting Equations~(\ref{eqn:two_cond}) and (\ref{eqn:req_cond}), 
it is seen that the left size of Equation~(\ref{eqn:req_cond}) is 
equal to $P(\vecx)$
and its right side is equal to $P(\vecx\,|\,\bbc_i)$.
Thus, the proposition follows from Equation~(\ref{eqn:req_cond}). \QED

\medskip

\section{Additional Material for Section~\ref{sec:utility_ext}}
\label{app:sec:utility_ext}

\medskip

\subsection{Statement and Proof of Lemma~\ref{lem:exp_total_util}}

\medskip

\underline{Statement of Lemma~\ref{lem:exp_total_util}:}~
Let $U_k$ denote the utility assigned to class $k$, $1 \leq k \leq K$.
Suppose the instances covered by a PSV combination \vecx{} are 
distributed uniformly randomly over the $K$ classes.
Then the total expected utility of the instances covered by \vecx{} is
$(N_{\vecx}/K) \sum_{k=1}^{K} U_k$, 
where $N_{\vecx}$ is the number of instances covered by \vecx{}
in the population.

\smallskip

\noindent
\textbf{Proof:}~ 
Let $\ell = N_{\vecx}$ and $M = \{w_1, w_2, \ldots, w_{\ell}\}$
be the set of all instances in the population covered by \vecx.
Let $h_i$ be the random variable that gives
the utility of $w_i$ when the instances in $M$ are distributed
uniformly randomly across the $K$ classes, $1 \leq i \leq \ell$.
Thus, the random variable $H = \sum_{i=1}^{\ell} h_i$ gives the total
utility of the instances in $M$.
By linearity of expectation \cite{MU-2005}, we have $E[H]$ = $\sum_{i=1}^{\ell} E[h_i]$. 
To find $E[h_i]$, we note that the probability that $w_i$ gets assigned to any
specific class $k$ is $1/K$ and the corresponding utility is $U_k$.
Therefore, $E[h_i] = \sum_{k=1}^{K} U_k/K$ = $(1/K) \sum_{k=1}^{K} U_k$.
Hence, $E[H]$ = $\sum_{i=1}^{\ell} E[h_i]$ = $(\ell/K) \sum_{k=1}^{K} U_k$.
Since $\ell = N_{\vecx}$, the lemma follows. \QED

\medskip

\subsection{Example of Unfairness Using Utility-Weighted Unfairness}
\label{app:sse:unfairness_ex}

We mentioned in Section~\ref{sec:utility_ext} 
that while our count-based formulation (Problem~\ref{prob:beta})
may not reveal unfairness, the utility-based formulation
(Problem~\ref{prob:Wutility}) can reveal unfairness.
Here, we present an example to illustrate this.

\medskip

\noindent
\textbf{Example:}~ Suppose $S$ is a set with 32 instances and suppose 
8 instances of $S$ are covered by a PSV combination $\vecx${}.
Thus, $P(\vecx) ~=~ 8/32 ~=~ 1/4$ and $N_{\vecx} ~=~ 8$.
Assume further that $S$ is partitioned into two classes $\bbc_1$ and $\bbc_2$
such that the following conditions hold: 
\begin{description}
\item{(i)} $|\bbc_1| = 24$ and 6 instances of $\bbc_1$ are covered by \vecx.
\item{(ii)} $|\bbc_2| = 8$ and 2 instances of $\bbc_1$ are covered by \vecx.
\end{description}
We note that $P(\vecx\,|\,\bbc_1) ~=~ P(\vecx\,|\,\bbc_2) ~=~ 1/4$.
In other words, for $i = 1, 2$, 
$P(\vecx\,|\,\bbc_i) ~=~ P(\vecx\,|\,\neg \bbc_i)$.  
Hence, by the formulation of count-based unfairness (Problem~\ref{prob:beta}), 
this classification is fair.

\smallskip

Now, suppose we assign the utility value $U_1 = 1$ and $U_2 = 4$ for the
two classes $\bbc_1$ and $\bbc_2$ respectively.
For these utility values, the values of the two expressions used
in the formulation of Problem~\ref{prob:Wutility} are as follows.
\begin{description}
\item{(i)} The value of the expression 
$|\bbc_1|P(\vecx{} |\bbc_k)\,U_1 + |\bbc_2|P(\vecx{} |\bbc_2)\,U_2$ is 
given by $24 \times (1/4) \times 1 + 8 \times (1/4) \times 4$ ~=~ 14.

\item{(ii)} The value of the expression $(N_{\vecx}/2) (U_1 + U_2)$ is
given by $(8/2) \times (1+4) ~=~ 20$.
\end{description}
Thus, we have utility values $U_1 = 1$ and $U_2 = 4$ such
that in the formulation of Problem~\ref{prob:Wutility}, 
$\alpha = 14$, $\beta = 20$ and $\alpha < \beta$.
Therefore, the utility-weighted fairness formulation points out
a possible unfairness situation while the count-based formulation
does not detect unfairness.


\section{Additional Material for Section~\ref{sec:complexity}}
\label{app:sec:addl_complexity}

\medskip

\subsection{Statement and Proof of Theorem~\ref{thm:ab_unfairness_hard}}

\medskip

\noindent
\underline{Statement of Theorem~\ref{thm:ab_unfairness_hard}:}
The \dugc{} problem is \cnp-complete even for two classes.

\medskip
  
\noindent
\textbf{Proof:}~
It is easy to see that \dugc{} is in \cnp{}
since given a subset $P'$ of PSVs one can efficiently
check that $P'$ covers at most $\alpha$ instances in \bbo{}
and at least $\beta$ instances in each of the other classes.

\medskip

To prove \cnp-hardness, we use a reduction from
the \textbf{Minimum Set Cover} (MSC) problem: given a universe 
$U = \{u_1, u_2, \ldots, u_n\}$, a collection 
$S$ = $\{S_1$, $S_2$, $\ldots$, $S_m\}$,
where each $S_j$ is a subset of $U$ ($1 \leq j \leq m$) and an integer $r \leq m$,
is there is a subcollection $S'$ of $S$ such that $|S'| \leq r$ and the union
of the sets in $S'$ is equal to $U$?
It is well known that MSC is \cnp-complete even when $r < n$
\cite{GJ-1979}.
The reduction from MSC to \dugc{} is as follows.
This reduction produces two classes, namely a target
class \bbo{} and a class \bbn.

\medskip

\noindent
\underline{Intuitive idea behind the reduction:}~ The target class \bbo{}
contains objects corresponding to the sets in the MSC problem.
The other class \bbn{} contains objects corresponding to
the universe in the MSC problem.
Each set in the MSC problem also represents a PSV.
The reduction specifies that the chosen combination of PSVs must cover
at most $r$ objects from \bbo{}
(to enforce the upper bound on the size of the solution
to MSC) and all $n$ objects in \bbn{} (to ensure that a set
collection that covers all the elements of $U$ can be obtained
from the chosen PSV combination).

\smallskip

The details of the reduction are as follows.

\smallskip

\begin{enumerate}[leftmargin=*,noitemsep,topsep=0pt]
\item The set of PSVs $\bbp{} = \{p_1, p_2, \ldots, p_m\}$
is in one-to-one correspondence with the collection
$S$ = $\{S_1$, $S_2$, $\ldots$, $S_m\}$.

\item We set $\beta = |U| = n$.
The class $\bbn{} = \{a_1, a_2, \ldots, a_{n}\}$ with $n$ instances
is in one-to-one correspondence 
with the universe $U = \{u_1, u_2, \ldots, u_n\}$.

\item Suppose the element $u_i$,~ $1 \leq i \leq n$,~ appears in subsets 
$S_{i_1}$, $S_{i_2}$, $\ldots$, $S_{i_t}$ for some $t \geq 1$. 
Then, for the instance $a_i \in \bbn$, $1 \leq i \leq n$, the PSVs
$p_{i_1}$, $p_{i_2}$, $\ldots$, $p_{i_t}$ have the value 1 and the
remaining PSVs have value 0.

\item We set $\alpha = r$ where $r$ is the bound on the number of sets
in the MSC instance.
The target class $\bbo{} = \{b_1, b_2, \ldots, b_m\}$ has 
$m$ instances. 
Since $r < n$ in the MSC problem, we satisfy the constraint
that $\alpha < \beta$ in the \dugc{} problem.

\item
For each instance $b_j \in \bbo$, where $1 \leq j \leq m$,
the PSV $p_j$ has the value 1 and the other PSVs 
have the value 0.
\end{enumerate}

\smallskip

\noindent
This completes our polynomial time reduction. 
We will now prove that there is a solution to the \dugc{} problem
iff there is a solution to the MSC problem.

\smallskip

Suppose $S' = \{S_{j_1}, S_{j_2}, \ldots, S_{j_{\ell}}\}$, where 
$\ell \leq r$, is a solution to the MSC problem. 
We first show that the subset $P' = \{p_{j_1}, p_{j_2}, \ldots, p_{j_{\ell}}\}$
covers $\beta = n$ instances in $\bbn$.
To see this, consider any instance $a_i \in \bbn$.
Since $S'$ is a solution to MSC, there is a set $S_{j_t} \in S'$
that covers the element $u_i \in U$ corresponding to $a_i$.
By our construction, the PSV $p_{j_t}$ has the value 1 for $a_i$ and
therefore $P'$ covers $a_i$.
Further, $P'$ covers $\ell \leq r = \alpha$ instances in $\bbo$
since each PSV in $P'$ covers exactly one instance in $\bbo$.
Thus, $P'$ is a solution to the \dugc{} problem.

\smallskip

Suppose $P' = \{p_{j_1}, p_{j_2}, \ldots, p_{j_{\ell}}\}$
is a solution to the \dugc{}  problem. 
If $|P'| = \ell > r = \alpha$, then again 
$P'$ would cover $\alpha+1$ or more instances in $\bbo$.
Therefore, $|P'| = \ell \leq r = \alpha$.
Let $S' = \{S_{j_1}, S_{j_2}, \ldots, S_{j_{\ell}}\}$ be
the subcollection of $S$ constructed from $P'$.
To see that $S'$ forms a solution to MSC, consider any element $u_i \in U$.
Since $P'$ is a solution to \dugc, there is a PSV,
say $p_{j_y} \in P'$, that covers $a_i \in \bbn$, 
the instance corresponding to $u_i \in U$.
By our construction of $S'$, the element $u_i$ is 
covered by the set $S_{j_y} \in S'$.
Thus, $S'$ forms a solution to the MSC problem, and this
completes our proof of Theorem~\ref{thm:ab_unfairness_hard}. \QED

\section{Additional Material for Section~\ref{sec:practice}}
\label{app:sec:addl_experiments}

\medskip

\subsection{The Congressional Districts in California, USA}
\label{app:sse:cong_dist}

\begin{figure}
\centering
\includegraphics[scale=0.16]{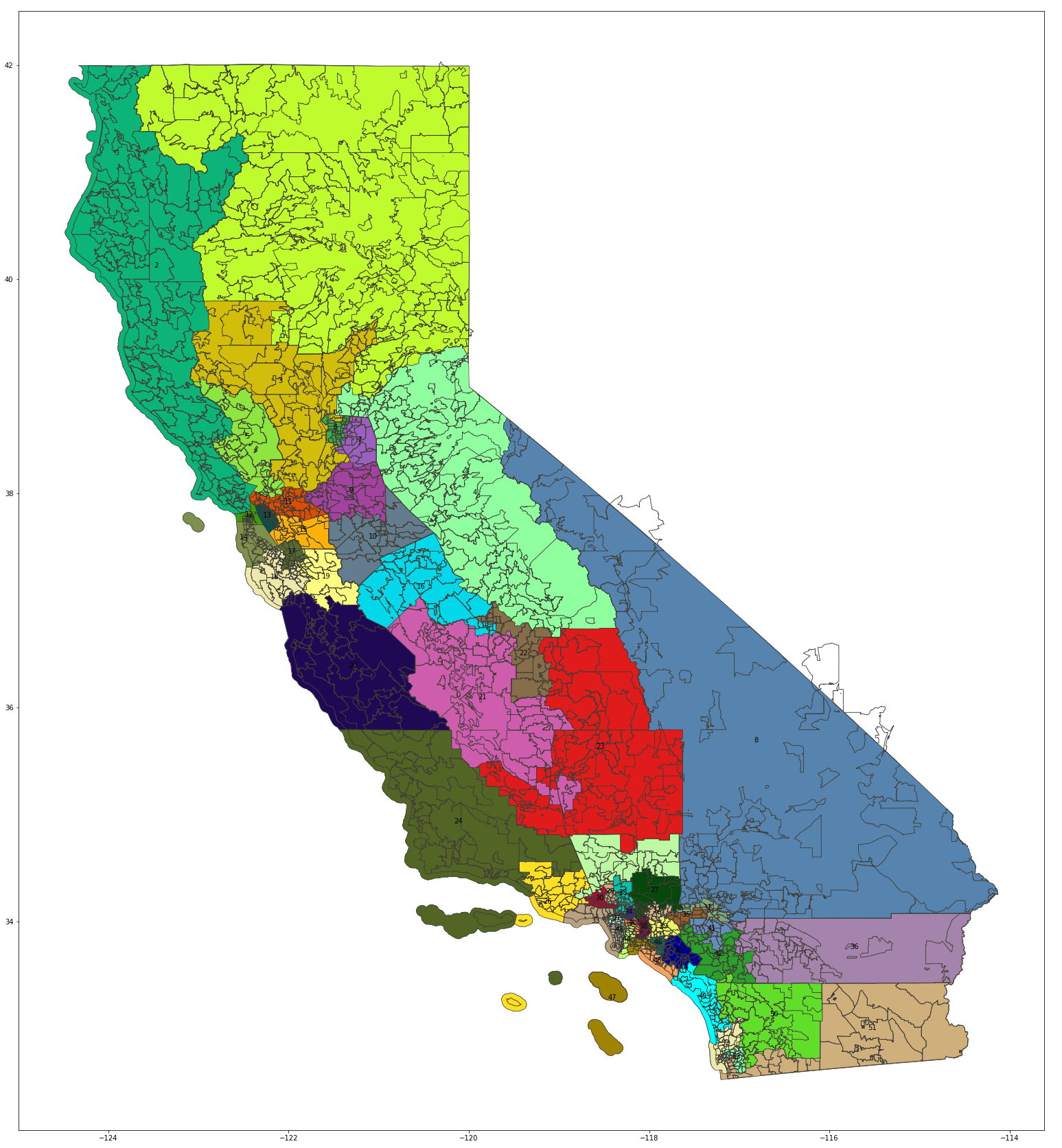}
\caption{The 53 California congressional districts (the classes) 
and the 1700+ ZCTA (Zip Code Tabulated Areas) that comprise them (the instances).}
\label{fig:CD}
\end{figure}

\clearpage

\subsection{Fair Reading Sources}

\begin{table}
{\small
\begin{center}
\begin{tabular}{|c|}
\hline
Wed Apr 01 22:39:24 +0000 2015 \\ Blood test for Down's syndrome hailed \\
        {\small \textsf{http://bbc.in/1BO3eWQ}}~~
        {\small \textsf{http://bbc.in/1ChTANp}} \\ \hline
Wed Apr 08 18:05:28 +0000 2015 \\ New approach against HIV `promising' \\ 
        {\small \textsf{http://bbc.in/1E6jAjt}} \\ \hline
Thu Apr 09 01:31:50 +0000 2015 \\ Breast cancer risk test devised \\ 
        {\small \textsf{http://bbc.in/1CimpJF}} \\ \hline
Tue Apr 07 00:04:09 +0000 2015 \\ Why strenuous runs may not be so bad after all \\ 
        {\small \textsf{http://bbc.in/1Ceq0Y7}} \\ \hline
Mon Apr 06 07:46:44 +0000 2015 \\ VIDEO: Health surcharge for non-EU patients \\ 
        {\small \textsf{http://bbc.in/1C5Mlbk}} \\ \hline
\end{tabular}
\medskip
\end{center}
}
\caption{ 
This table shows a few examples of articles on health from the BBC health care
website. The first article would be tagged as being on handicapped,
the third about women and the fifth about poverty. 
Note that the tagging is based on an article's content and not its title.}
\label{tab:exfeeds}
\end{table}

\end{document}